\documentclass[letterpaper, 10 pt, conference]{ieeeconf}

\IEEEoverridecommandlockouts                              

\usepackage{graphicx} 
\usepackage{times} 
\usepackage{amsmath} 
\usepackage{amssymb}  
\usepackage{bm}
\graphicspath{{./Figures/}}
\usepackage[caption=false,font=footnotesize]{subfig}
\usepackage{braket}
\usepackage{floatrow}
\usepackage{tikz}
\usepackage{floatrow}
\usepackage{svg}
\usepackage[hidelinks,pdfpagemode=UseNone]{hyperref}
\usepackage{algorithm}
\usepackage[noEnd=false]{algpseudocodex}

\usepackage{booktabs,ragged2e}
\usepackage[flushleft]{threeparttable}

\usepackage{xcolor}

\definecolor{orange}{rgb}{1.0, 0.22, 0.0}

\title{\LARGE \bf
Motion Planning for Object Manipulation by Edge-Rolling}

\author{Maede Boroji\textsuperscript{\textdagger}{\thanks{\textsuperscript{\textdagger}
These authors contributed equally.}}, Vahid Danesh\textsuperscript{\textdagger}, Imin Kao, Amin Fakhari
\thanks{The authors are with the Department of Mechanical Engineering, Stony Brook University, Stony Brook, NY 11794, USA, {\tt\small \{maede.boroji, vahid.danesh, imin.kao, amin.fakhari\}@stonybrook.edu}.
}
}

\begin{document}

\maketitle
\thispagestyle{empty}
\pagestyle{empty}

\begin{abstract}
A common way to manipulate heavy objects is to maintain at least one point of the object in contact with the environment during the manipulation. When the object has a cylindrical shape or, in general, a curved edge, not only sliding and pivoting motions but also rolling the object along the edge can effectively satisfy this condition. \textit{Edge-rolling} offers several advantages in terms of efficiency and maneuverability. This paper aims to develop a novel approach for approximating the prehensile edge-rolling motion on any path by a sequence of constant screw displacements, leveraging the principles of screw theory. Based on this approach, we proposed an algorithmic method for task-space-based path generation of object manipulation between two given configurations using a sequence of rolling and pivoting motions. The method is based on an optimization algorithm that takes into account the joint limitations of the robot. To validate our approach, we conducted experiments to manipulate a cylinder along linear and curved paths using the Franka Emika Panda manipulator.

\textit{Video}--- \url{https://youtu.be/MX1-MAR9ubc}


\end{abstract}

\section{Introduction}

A common task performed by robotic manipulators involves pick-and-place operations, wherein the manipulator grasps an object, lifts it off the environment, and places it in a designated location. However, this scenario becomes ineffective when the manipulator must handle a heavy object. An object is considered heavy if its weight exceeds the joint torque limits of the manipulator (or its gripper) required to lift it off the environment. It is still possible to manipulate heavy objects with a given manipulator if we maintain at least one point of the object in contact with the environment, allowing for full or partial compensation of the object's weight. The three primitive motions used to manipulate an object while maintaining contact with the environment are: (i) sliding or pushing on a vertex, edge, or face of the object, (ii) pivoting about an axis passing through a vertex, edge, or face of the object, and (iii) rolling on a curved edge or face of the object. By combining these primitive motions, it is possible to achieve all other types of motions.

Sliding or pushing motions can be advantageous in various scenarios, however, they will be infeasible in situations involving rough surfaces (high friction coefficient) or steps. In these situations, pivoting motions are more efficient as we investigated in our previous research~\cite{MotionForcePlanning2021}. When dealing with heavy cylindrical objects, relying solely on sliding or pivoting may not provide the most effective solution. For example, a common method for a human to move a heavy carpet roll or cylindrical container (e.g., a barrel, gas tank, or steel drum), is to tilt it onto its side and then, roll it on the edge. Although rolling on the face may be feasible in some situations, rolling on the edge offers a distinct advantage by requiring significantly smaller footprints and higher maneuverability.

\begin{figure}[t]
    \centering
    \includegraphics[width=0.9\textwidth]{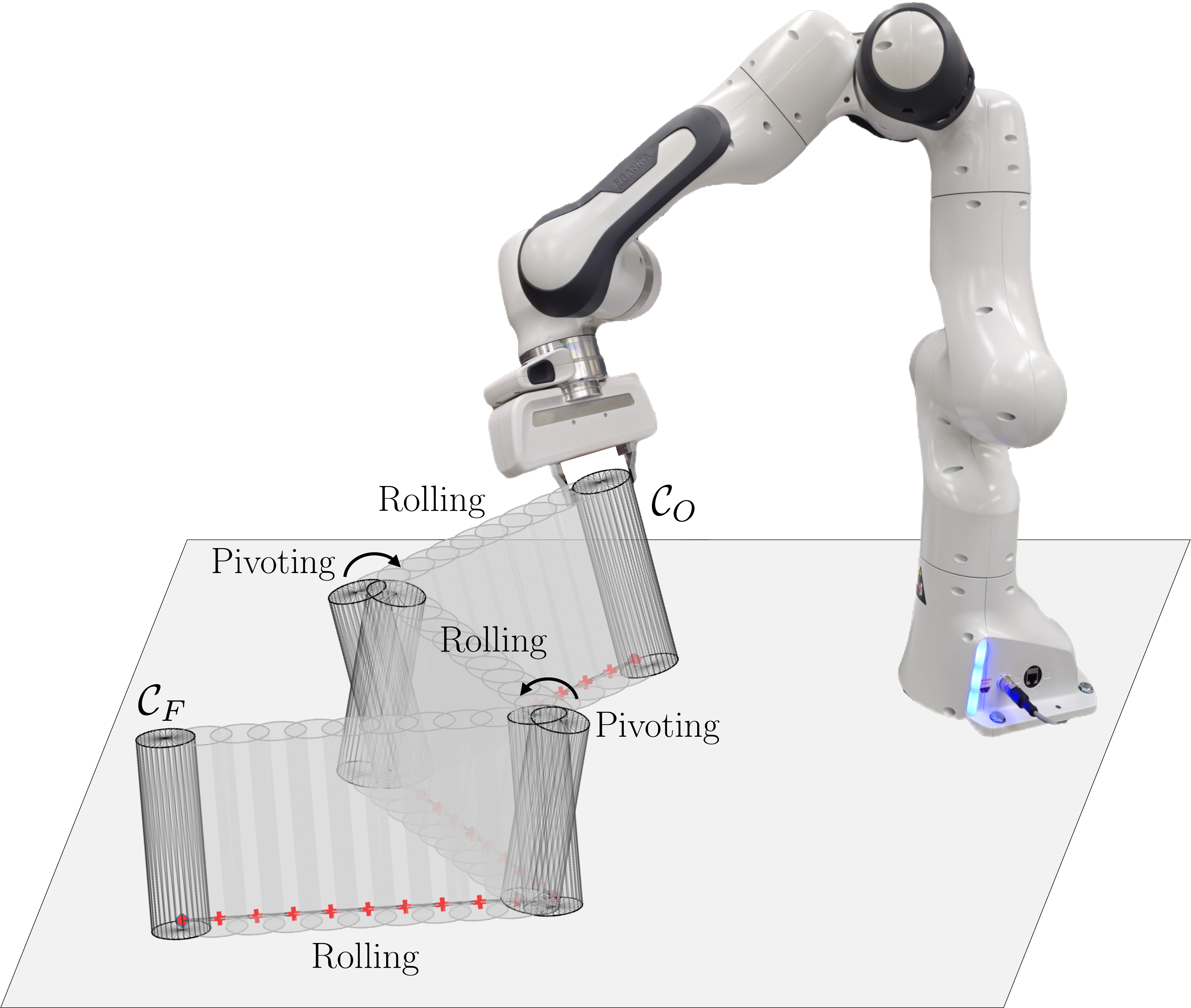}
    \caption{Schematic sketch for the manipulation of a cylinder through a sequence of pivoting and rolling motions, transitioning it from an initial configuration $\mathcal{C}_O$ to a final configuration $\mathcal{C}_F$.}
    \label{fig:FirstPage_Figure}
\end{figure}

In this paper, we focus on the problem of planning motion for prehensile manipulation of a cylindrical object, involving a combination of rolling and pivoting motions, from an initial configuration $\mathcal{C}_O$ to a final configuration $\mathcal{C}_F$ as shown in Fig.~\ref{fig:FirstPage_Figure}. Here, prehensile manipulation refers to the absence of slippage at the interfaces between the object and the manipulators. The key challenge in motion planning is that in this motion a point of the object (instantaneously) remains in contact with the environment, forming a manipulator-object-environment closed-chain constraint. Moreover, the rolling motion imposes a nonholonomic constraint.



\subsection{Related Work}


A method for whole-body pushing manipulation with contact posture planning for humanoid robots using various contact points depending on the pushing force, was proposed by Murooka \textit{et. al.}~\cite{Murooka_WholeBody_Pushing}. Polverini \textit{et. al.}~\cite{Polverini_Pushing} presented a control architecture for torque-controlled platforms that enables multi-contact loco-manipulation actions, to perform the pushing task. The task of pushing cylinder-shaped objects to reach a desired goal pose, was studied in~\cite{pushing_Cylinder}. Pivoting was ﬁrst introduced
by Aiyama \textit{et. al.}~\cite{Aiyama_Pivoting} as a new method of non-prehensile graspless manipulation. Yoshida \textit{et. al.}~\cite{Yoshida_Pivoting,Yoshida_MotionPlanning,Pivoting_based_manipulation} conducted a controllability analysis of pivoting manipulation of polyhedral objects in humanoid robots, showing the ability to accurately displace bulky objects to desired pose on a plane, introducing a steering method that converts nonholonomic paths into a sequence of pivoting operations. The analysis of pivoting and tilting as a form of extrinsic regrasping was studied in~\cite{openloop_Pivoting}.
The concept of manipulating an object while maintaining external contact with the object has been referred to as `shared grasping' in~\cite{SharedGrasping}.
In our previous works~\cite{ForceSynthesis2020,GraspMetric2021,MotionForcePlanning2021}, we have studied the motion and force planning of pivoting heavy objects. Specifically, we have introduced the concept of \textit{object gaiting} as an algorithmic strategy to manipulate heavy objects by a sequence of pivoting motions. 

Transporting cylindrical objects with a focus on non-prehensile grasp planning and the design of suitable end-effectors for stable grasp was introduced by Specian \textit{et. al.}~\cite{Edge_Rolling2018}. Picking and placing objects with different surface and shape conditions, including cylindrical objects, was performed using a customized end-effector featuring a curved finger, by Mucchiani \textit{et. al.}~\cite{Cylinder_Picking}. Chavan-Dafle \textit{et. al.}~\cite{Prehensile_pushing} demonstrated that cylindrical objects can be manipulated over their curved face when held loosely by fingers (non-prehensile grasp) and this manipulation can involve a combination of rolling and sliding motions. However rolling on a curved face can pose difficulties due to the potentially large footprints involved, particularly in cluttered environments.

Previous studies have extensively investigated motions involving sliding/pushing and pivoting.
Therefore, there exists a noticeable lack of exploration concerning the fundamental aspects of prehensile edge-rolling motion. This study aims to fill this gap by investigating the underlying principles and developing algorithmic methods for edge-rolling motion.


\subsection{Contributions}
We present an approach to approximate the rolling motion by a sequence of constant screw motions. This approach can mathematically model edge-rolling on both straight-line and general curved paths. By using the screw displacement, we can, not only implicitly satisfy the closed-chain constraint during the motion but also achieve a unified method of representing the primitive motions of sliding, pivoting, and rolling based on screw theory. We also developed a method to edge-roll the object between any two arbitrary configurations when a direct straight-line motion between them is not feasible, due to the joint limits of the manipulators. This method is based on the integration of the pivoting motion represented by ScLERP, the proposed screw-based representation of rolling, and an optimization algorithm. To the best of our knowledge, this is the first study on the prehensile manipulation of curved-edge objects by rolling. 



\section{Preliminaries}
Dual quaternions~\cite{kavan2006dual} offer an efficient and convenient representation of screw motions due to their unified representation of translation and rotation, compactness, interpolation capabilities, ease of differentiation, and compatibility with screw theory principles. Therefore, in this section, we will provide a brief overview of dual quaternions.

A dual number is defined as $\bar{d} = a + \epsilon b$, where $a$ and $b$ are real numbers (or more generally, elements of a field), and $\epsilon$ is a dual unit with $\epsilon^2 = 0$, $\epsilon \neq 0$. In this expression, $a$ and $b$ are referred to as the real and dual parts of $\bar{d}$, respectively.
Dual quaternion algebra is an extension of the dual-number theory by Clifford ~\cite{Clifford1871PreliminarySO} that can provide an elegant approach to solving a variety of complex problems and a smooth and constant interpolation between two rigid transformations.
A dual quaternion, denoted as $\sigma$, takes the form $\sigma = p + \epsilon q$, where $p \in \mathbb{H}$ and $q \in \mathbb{H}$ are quaternions defined as $p = p_0 + \boldsymbol{p} = p_0 + p_1 \mathrm{i} + p_2 \mathrm{j} + p_3 \mathrm{k}=(p_0, \boldsymbol{p})$,
$q = q_0 + \boldsymbol{q} = q_0 + q_1 \mathrm{i} + q_2 \mathrm{j} + q_3 \mathrm{k}=(q_0, \boldsymbol{q})$.
Here, $p$ and $q$ represent the real and dual parts of $\sigma$, respectively. Alternatively, $\sigma$ can be viewed as an 8-tuple $\sigma = (p_0, p_1, p_2, p_3, q_0, q_1, q_2, q_3)$.
The addition and multiplication operations for two dual quaternions, $\sigma_1 = p_1 + \epsilon q_1$ and $\sigma_2 = p_2 + \epsilon q_2$, are performed as follows: 
$\sigma_1 + \sigma_2 = (p_1 + p_2) + \epsilon (q_1 + q_2)$ and $\sigma_1 \cdot \sigma_2 = (p_1 \cdot p_2) + \epsilon (p_1 \cdot q_2 + q_1 \cdot p_2)$.
A dual quaternion $\sigma = p + \epsilon q$ is considered a unit if
it satisfies these two conditions $p_0^2 + p_1^2 + p_2^2 + p_3^2 = 1$ and $p_0q_0 + p_1q_1 + p_2q_2 + p_3q_3 = 0$.

Unit dual quaternions are commonly used to represent rigid body transformation (rotation and translation) in the group $SE(3) = \mathbb{R}^3 \times SO(3) = (\boldsymbol{p},\boldsymbol{R})$, where $\boldsymbol{p} \in \mathbb{R}^3$ and $\boldsymbol{R} \in SO(3)$.
Let $Q_R = (\cos\frac{\theta}{2}, \boldsymbol{u}\sin\frac{\theta}{2}) \in S^3$ be a unit quaternion representation of pure rotation of the body about a unit vector $\boldsymbol{u}$ by an angle $\theta$, and $Q_p = (0,\boldsymbol{p})\in S^3$ be a unit quaternion representation of pure translation of the body along vector $\boldsymbol{p}$. Therefore, the unit dual quaternion representation of the rigid body transformation is $D_T = Q_R + \epsilon \frac{1}{2} Q_p Q_R $.
If the transformation is a pure rotation, $\boldsymbol{p} = \boldsymbol{0}$ and we have $D_T = Q_R + 0 \epsilon$. If the transformation is a pure translation, $\theta = 0$, and we end up with $D_T = 1 + \epsilon \frac{1}{2} Q_p$.


\section{Problem Statement}
\begin{figure}[!htbp]
    \centering
    \includegraphics[width=0.9\textwidth]{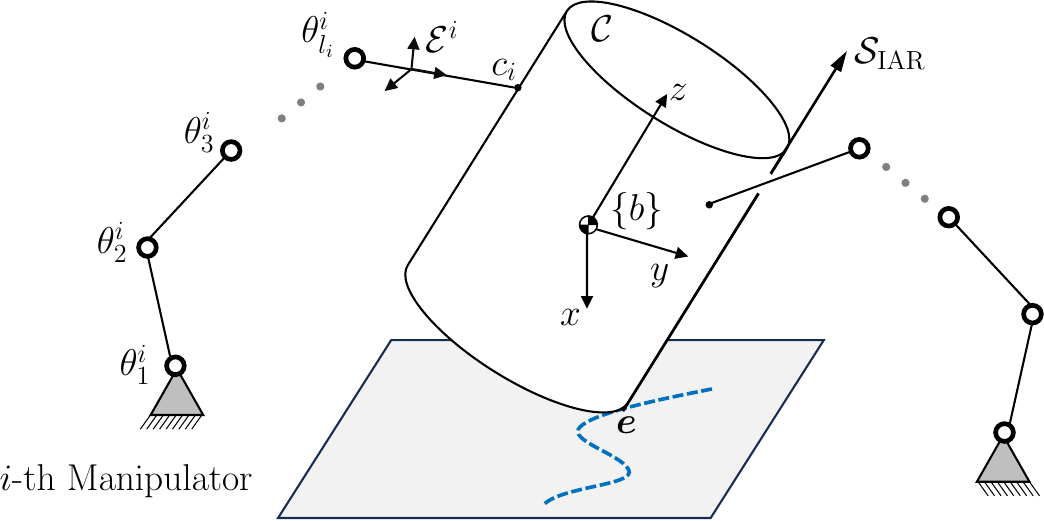}
    \caption{A cylindrical-shape object being rolled by $n$ manipulators.}
    \label{fig:ProblemStatement}
\end{figure}

Consider a scenario where an object with a circular-shaped edge (e.g., a cylinder or cone)
in contact with the environment is being manipulated by $n$ $l_i$-DOF ($i=1,...,n$) manipulators or fingers. The goal is to move the object from an initial pose $\mathcal{C}_O \in SE(3)$ to a final pose $\mathcal{C}_F \in SE(3)$ while ensuring continuous contact of its circular edge with the environment (Fig.~\ref{fig:ProblemStatement}). We assume that (i) the motion is quasi-static, (ii) both the initial pose $\mathcal{C}_O$ and the final pose $\mathcal{C}_F$ lie within the robot's workspace, (iii) the position of the contact points between the manipulators and the object, denoted as $c_i$, are provided, and (iv) no slippage is occurring at the object-environment and object-manipulators contact interfaces.

Our solution involves modeling the rolling motion by a sequence of constant screw displacements, where the screw axis always passes through the object-environment contact point. In Sec.~\ref{sec:Edge-Rolling}, we discuss rolling on a straight line, where the screw axis is always parallel to the cylinder's axis of symmetry. In Sec.~\ref{sec:Pivoting}, we generalize the method for rolling on a curve, where a screw axis perpendicular to the environment supporting plane is also involved.

\section{Edge-Rolling on a Straight-Line Path} 
\label{sec:Edge-Rolling}
In edge-rolling motion, an object always maintains a point $\boldsymbol{e}$ of its edge in contact with a supporting surface and instantaneously rotates about an axis that passes through this point. This axis is called the Instantaneous Axis of Rotation (IAR) as shown in Fig.~\ref{fig:ProblemStatement}. Thus, the motion of the object is instantaneously a pure rotation about an axis and restricted to $SO(3)$, which is a subgroup of $SE(3)$. We represent this axis as a screw axis $\mathcal{S}_\mathrm{IAR}$ in the context of screw displacement and this is the basis of our methodology for the edge-rolling motion.

In general, any rigid body displacement can be expressed as a rotation $\theta \in \mathbb{R}$ about a screw axis $\mathcal{S}$ followed by a translation $d \in \mathbb{R}$ along the axis. By representing the screw axis $\mathcal{S}$ by a unit vector $\boldsymbol{u} \in \mathbb{R}^3$ along the axis and an arbitrary point $\boldsymbol{r} \in \mathbb{R}^3$ on the axis, the screw parameters, using Plücker coordinates, are defined as $(\theta,d,\boldsymbol{u},\boldsymbol{m})$, where $\boldsymbol{m}= \boldsymbol{r} \times \boldsymbol{u} \in \mathbb{R}^3$. Therefore, the screw displacements can be efficiently expressed by the dual quaternions as $ D_T = Q_R + \epsilon \frac{1}{2} Q_P Q_R =( \cos\frac{\theta}{2} + \boldsymbol{u}\sin\frac{\theta}{2}) + \epsilon\left(-\frac{d}{2}\sin\frac{\theta}{2}+\sin\frac{\theta}{2}\boldsymbol{m}+\frac{d}{2}\cos\frac{\theta}{2}\boldsymbol{u}\right)$. In a \textit{constant} screw motion/displacement, $\boldsymbol{u}$ and $\boldsymbol{m}$ remains constant and only $\theta$ and/or $d$ changes.

\begin{figure}[!htbp]
    \centering
    \subfloat[]{\includegraphics[scale=0.42]{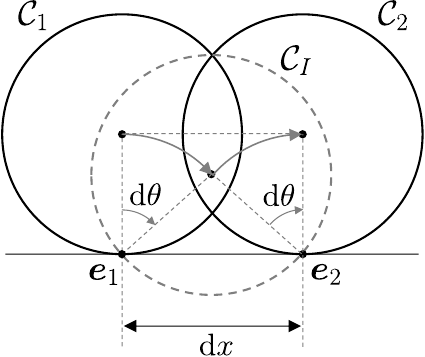}} \qquad 
    \subfloat[]{\includegraphics[scale=0.42]{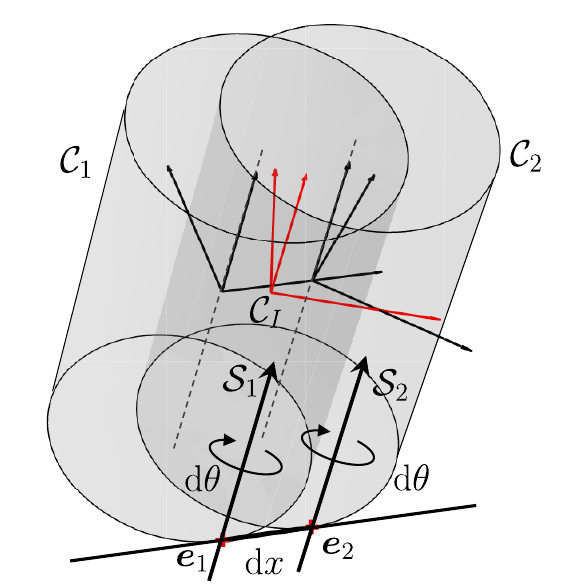}}
    \caption{ Moving from configuration $\mathcal{C}_1$ to configuration $\mathcal{C}_2$ of an element $\mathrm{d}x$ using an intermediate configuration $\mathcal{C}_I$ and two screw displacements, (a) demonstration on a 2D circle, (b) demonstration on a 3D cylinder.
    }
    \label{fig:figure_Mthod1}
\end{figure}

Our motion planning algorithm for edge-rolling can be more readily elucidated within the confines of a two-dimensional (2D) scenario. Assume that a 2D circle with a known radius $R$ is rolling along a straight-line path as shown in Fig.~\ref{fig:figure_Mthod1}. In this case, the IAR passes through the contact point, which is also known as the instantaneous center of rotation (ICR), toward the $z$-axis. Let $\mathrm{d}x$ be an element of the path between two points $\boldsymbol{e}_1$ and $\boldsymbol{e}_2$, and $\mathcal{C}_1$ and $\mathcal{C}_2$ corresponding to configurations where the contact point of the circle coincides with $\boldsymbol{e}_1$ and $\boldsymbol{e}_2$, respectively. By defining an intermediate configuration $\mathcal{C}_I$ such that it passes through the points $\boldsymbol{e}_1$ and $\boldsymbol{e}_2$, we can move the circle from $\mathcal{C}_1$ to $\mathcal{C}_2$ using two constant screw motions. In the first constant screw motion, the circle at the configuration $\mathcal{C}_1$ rotates about the screw axis $\mathcal{S}_1$ (that passes through the point $\boldsymbol{e}_1$) by $\mathrm{d}\theta$ to achieve the penetrated intermediate configuration $\mathcal{C}_I$, and in the second constant screw motion, the circle at the intermediate configuration $\mathcal{C}_I$ rotates about the screw axis $\mathcal{S}_2$ (that passes through the point $\boldsymbol{e}_2$) by the same $\mathrm{d}\theta$ to achieve the configuration $\mathcal{C}_2$. Therefore, for each element $\mathrm{d}x$ along the path we need to use two constant screw motions, and as $\mathrm{d}x$ decreases the motion gets closer to pure rolling (with zero slippage). The configuration $\mathcal{C}_2$ is now used as the initial configuration of the next element along a straight-line path. Using the geometry of the motion, $\mathrm{d}x = 2R\sin \mathrm{d}\theta$ and therefore,  ${\mathrm{d}\theta} = \sin^{-1}\left(\frac{\mathrm{d}x}{2R}\right)$.

All the non-intermediate configurations, e.g., $\mathcal{C}_1$ and $\mathcal{C}_2$, which are non-penetrated, are used for the motion planning of the manipulators. Therefore, we can compute all desired configurations corresponding to an approximated edge-rolling motion, without penetrating the supporting plane.

The concept of the rolling motion for a two-dimensional circle can be readily employed for the edge-rolling motion of a three-dimensional cylindrical object along a designated discretized path. As shown in Fig.~\ref{fig:figure_Mthod1}, for each element of the path, the cylinder moves from $\mathcal{C}_1$ to $\mathcal{C}_2$ using two constant screw motions about the screw axes $\mathcal{S}_1$ and $\mathcal{S}_2$ that passes through the point contacts.

In this section, we described our methodology for rolling the object along a straight-line path. To extend the method for rolling along a curved path, we need to first explain the pivoting motion.

\section{Edge-Rolling on a Curved Path}
\label{sec:Pivoting}

\begin{figure}[!htbp]
    \centering
    \includegraphics[scale=0.5]{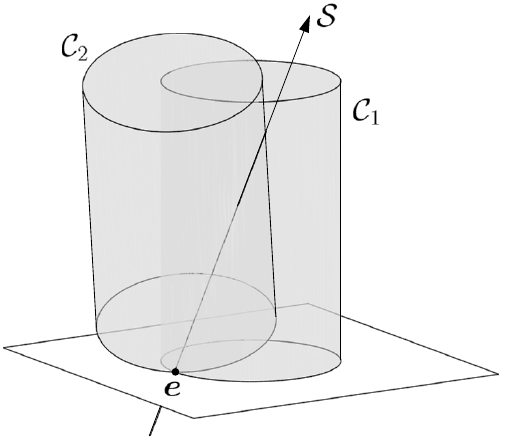}
    \caption{ScLERP between two given configurations $\mathcal{C}_1$ and $\mathcal{C}_2$ of a cylinder.}
    \label{fig:PivotingCylinder}
\end{figure}

\textbf{Pivoting Motion}: Pivoting involves moving an object while keeping contact with a support surface at a fixed point or line. In our previous paper~\cite{MotionForcePlanning2021}, we demonstrated that Screw Linear Interpolation (ScLERP) can be used to mathematically model this motion, where the screw axis passes through the fixed point or line. Therefore, when two configurations of a body have at least one point in common (Fig.~\ref{fig:PivotingCylinder}), the ScLERP can guarantee that this point remains fixed in all the intermediate configurations. Consequently, the body's non-penetration constraint is always fulfilled without explicitly enforcing it. The ScLERP provides the shortest straight line path in $SE(3)$ between two given configurations. Let $D_1$ and $D_2$ be the unit dual quaternion representation of configurations $\mathcal{C}_1$ and $\mathcal{C}_2$, respectively. The smooth path in $SE(3)$ provided by the ScLERP is derived by $D(s) = D_1D_{12}^{s}$, where $s \in [0, 1]$ is a scalar path parameter and $D_{12}$ is the transformation of $\mathcal{C}_2$ with respect to $\mathcal{C}_1$. $D_{12}^ {s}$ can be computed using $D_{12}^{s} = \left(\cos \frac{s \theta}{2}, \sin \frac{s\theta}{2}\boldsymbol{u}\right) + \epsilon \left(- \frac{s d}{2} \sin \frac{s \theta}{2}, \frac{s d}{2}\cos \frac{s \theta}{2} \boldsymbol{u} + \sin \frac{s \theta}{2} \boldsymbol{m}\right)$ after extracting the screw parameters $\boldsymbol{u}$, $\boldsymbol{m}$, $\theta$, $d$ from $D_{12}$. In this equation, $\theta = 0, \pi$ corresponds to pure translation between two poses when the screw axis is at infinity, and $d = 0$ corresponds to pure rotation.


When a cylindrical object is rolling along a curved path (e.g., polynomial or circular),
it needs to adjust its orientation to align with the direction of each segment of the path. Therefore, to achieve a rolling motion along a curved path, it is necessary to pivot the cylinder after each segment of rolling to align it with the subsequent element of the path.

\begin{figure}[!htbp]
    \centering
    \includegraphics[width=0.8\textwidth]{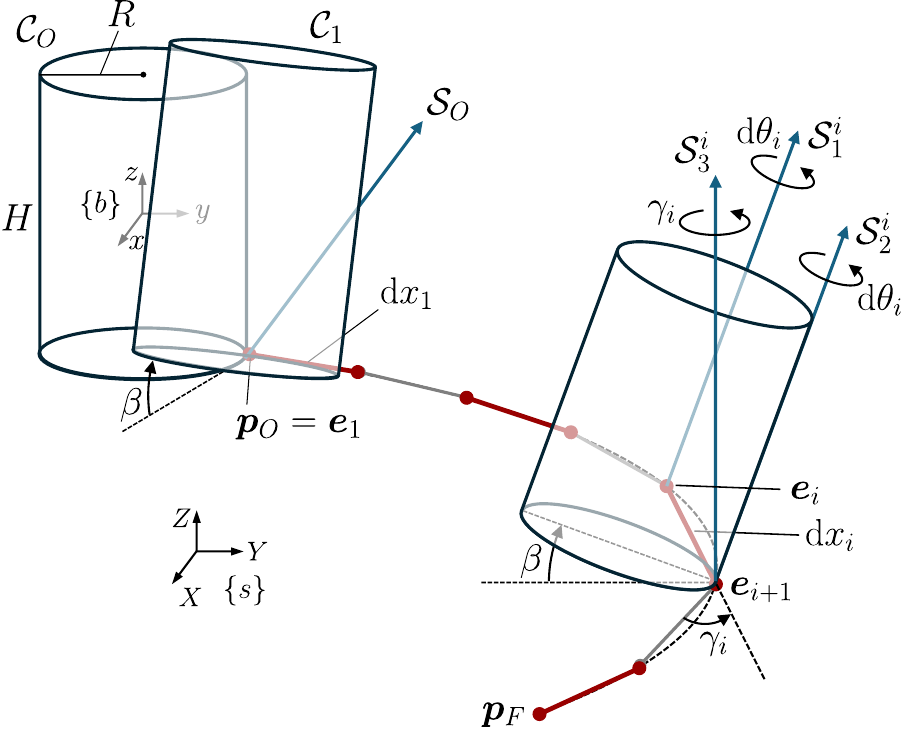}
    \caption{Edge-rolling of a cylinder on a discretized curved path. The screw axes $\mathcal{S}_1^i$ and $\mathcal{S}_2^i$ correspond to the constant screw displacements to roll the cylinder along the segment $dx_i$ and the screw axis $\mathcal{S}_3^i$ corresponds to a pivoting motion to align the cylinder with the next segment.}
    \label{fig:CylinderRolling_CurvedPath}
\end{figure}

Assume that we want to roll a cylinder that is initially at the configuration $\mathcal{C}_O$ on a curved path that intersects the circular footprint of the cylinder at point $\boldsymbol{e}_1$  as shown in Fig.~\ref{fig:CylinderRolling_CurvedPath}. After the discretization of the curved path by small linear segments $\mathrm{d}x_i$, to establish a point contact with the surface, the cylinder is first pivoted about point $\boldsymbol{e}_1$ to configuration $\mathcal{C}_1$ where the cylinder is aligned with the first segment of the path and has an angle $\beta$ with the supporting surface. This pivoting motion from $\mathcal{C}_O$ to $\mathcal{C}_1$ can be represented by a constant screw motion about a screw axis $\mathcal{S}_O$ using ScLERP (as explained in the first paragraph of Sec.~\ref{sec:Pivoting}). Once the object is aligned with the first segment, for each segment $\mathrm{d}x_i$ of the path, we need to use two constant screw motions about the screw axes $\mathcal{S}_1^i$ and $\mathcal{S}_2^i$ by $\mathrm{d}\theta_i$, as explained in Sec.~\ref{sec:Edge-Rolling}, and also a pivoting motion using ScLERP about the screw axis $\mathcal{S}_3^i$ (perpendicular to the supporting surface) by $\gamma_i$ (the angle between two consecutive elements) to align the object with the next segment. This pivoting motion ensures a smooth transition between segments. Once the object reaches the last segment of the path, the last pivoting is used to reach a given final configuration $\mathcal{C}_F$. Therefore, for edge-rolling a cylinder on a curved path, for each linear segment of the path we need two infinitesimal constant screw displacements and one pivoting motion. 

Depending on the magnitude of the angle $\gamma_i$, the pivoting motion can be a single screw displacement or ScLERP between two configurations. Moreover, to capture the curvature of the path with better approximation and minimize the slippage, it is necessary to choose the linear segments $\mathrm{d}x_i$ infinitesimally small. 


\begin{figure}[!htbp]
    \centering
    \subfloat[side view]{\includegraphics[width=0.5\textwidth]{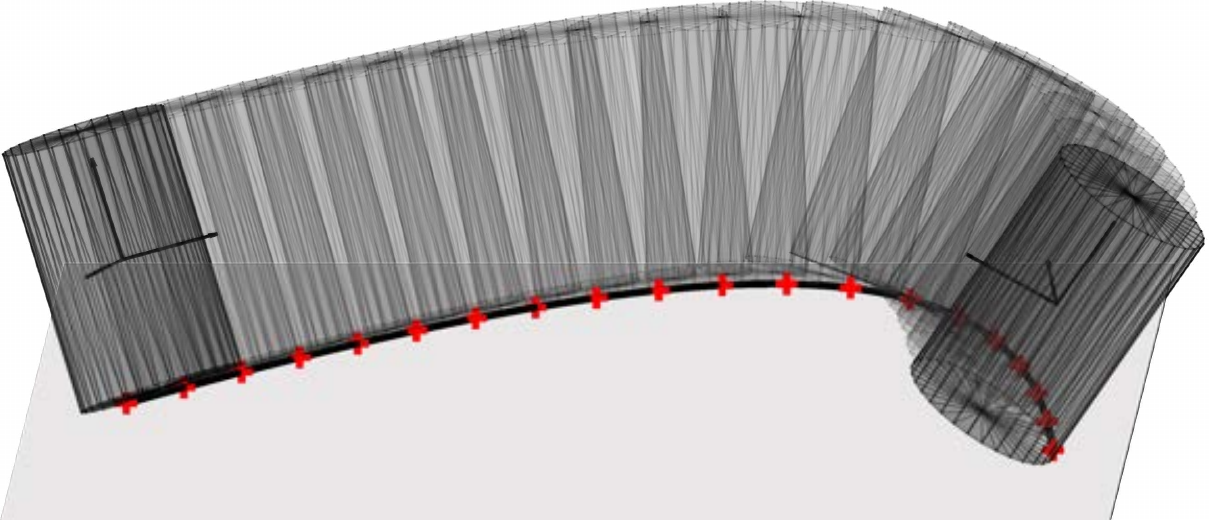}} \quad
    \subfloat[top view]{\includegraphics[width=0.3\textwidth]{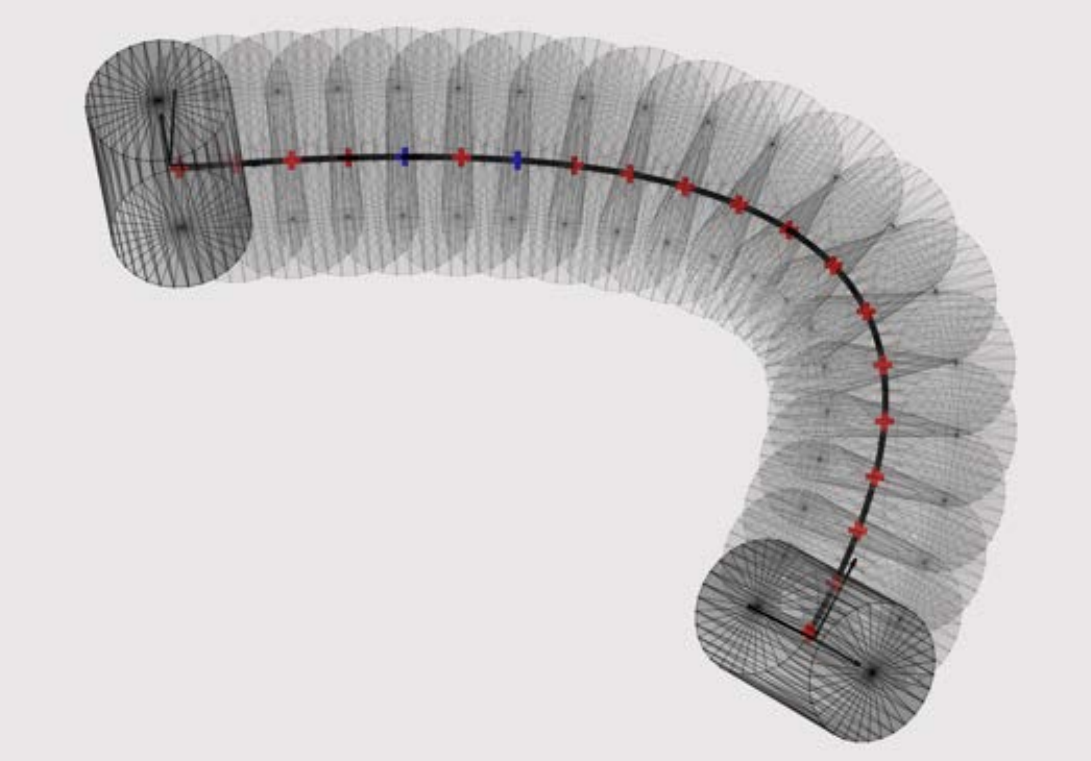}}
    \quad
    \subfloat[side view]{\includegraphics[width=0.5\textwidth]{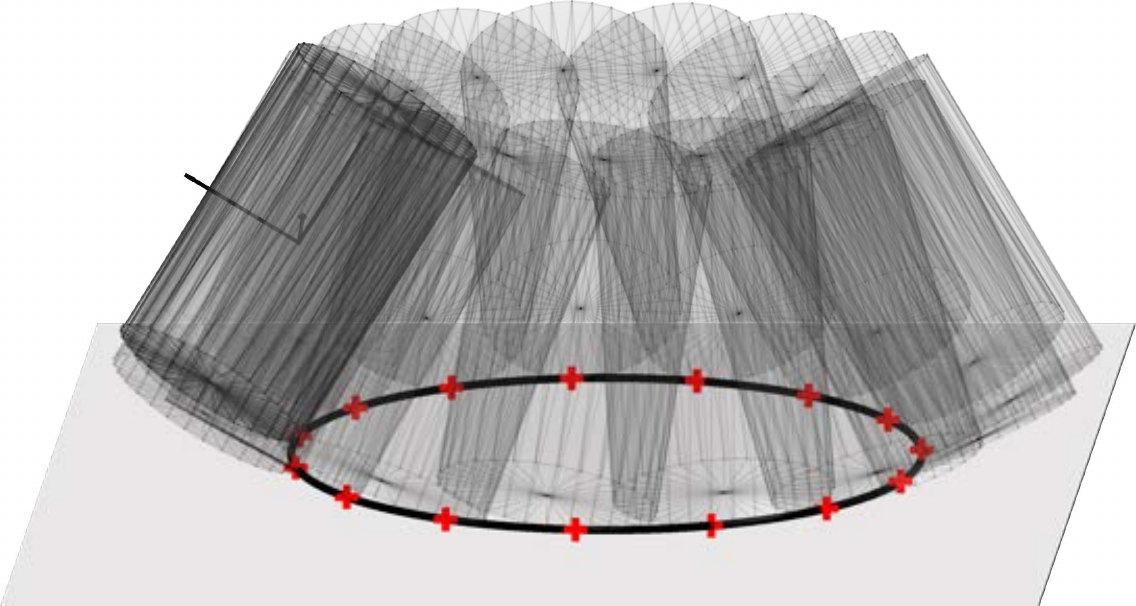}} \quad
    \subfloat[top view]{\includegraphics[width=0.3\textwidth]{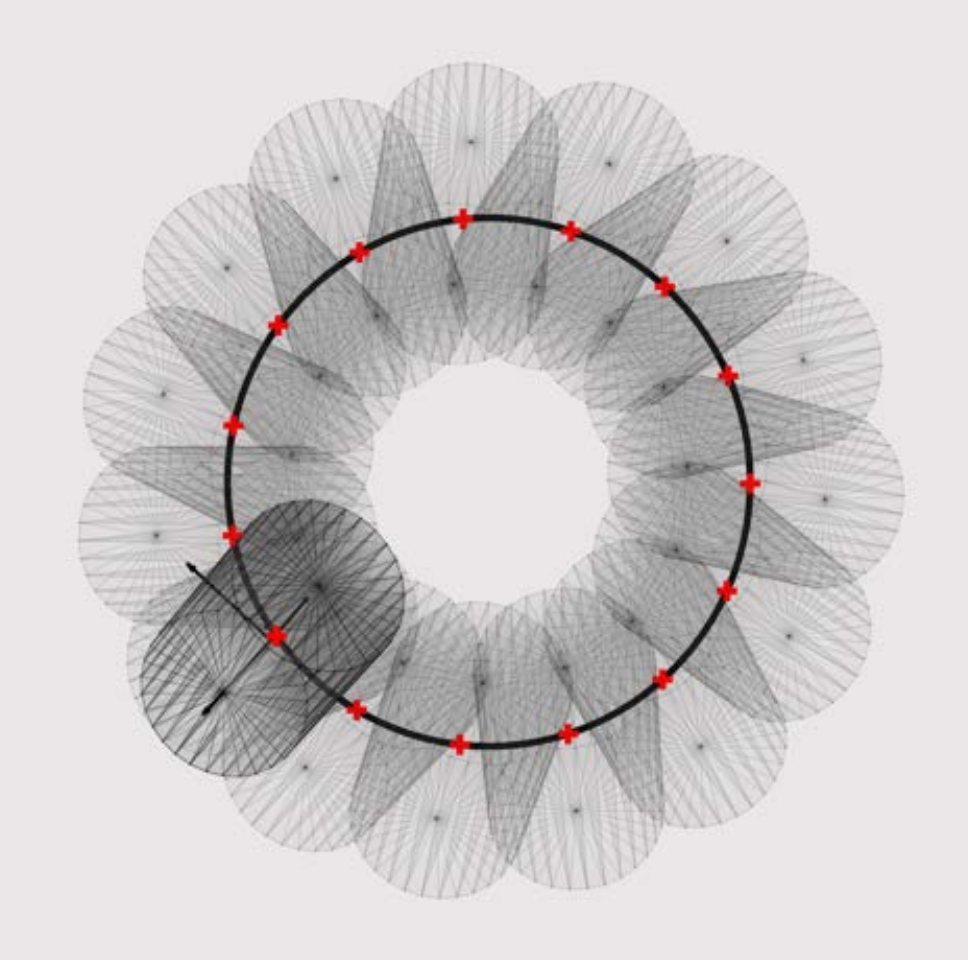}}
    \caption{Manipulation of the cylinder along (a, b) a second-order polynomial path and (c, d) along a circular path.}
    \label{fig:Cylinder_Polyline_Circular}
\end{figure}

Figure.~\ref{fig:Cylinder_Polyline_Circular} illustrates the edge-rolling of a cylinder along two paths; a second-order polynomial path (a,b) and a circular path (c,d), showcasing the versatility of the motion planning methodology.

The angle $\beta$ of the cylinder with the supporting surface (Fig.~\ref{fig:CylinderRolling_CurvedPath}) can remain constant or change along the path. To reduce the moment of the gravitational force (weight) of the body about the contact point $\boldsymbol{e}$ and enhance stability in the presence of external disturbance wrenches, one can align the center of mass directly above the contact point to ensure that the gravitational force passes through the contact point. The tilting angle $\beta$ for a cylinder with uniform mass, radius $R$, and height $H$ is computed as $\beta = \frac{\pi}{2} - \arctan\left(\frac{H}{2R}\right)$.


In this section, we assumed that the object's shape is cylindrical and the supporting surface is flat. However, this technique can be extended to any object with a curved edge and also a wide range of supporting surfaces, including non-flat surfaces, inclined surfaces, and uneven terrains.

For prehensile pushing or sliding an object along a curved path, we can also use a similar method. Instead of employing two screw displacements about the screw axes $\mathcal{S}_1^i$ and $\mathcal{S}_2^i$ for each segment, we use a single screw displacement (with pure translation) along the segment. The pivoting motion at the end of each segment about the screw axis $\mathcal{S}_3^i$ remains unchanged.

Therefore, we can model and implement all the primitive motions, i.e., sliding, rolling, and pivoting, where at least one point of the body maintains its contact with the environment, using a unified principle based on screw theory. This enables us to readily combine these methods and create a general motion that incorporates all types of primitive motions, providing a versatile and comprehensive approach to motion planning. The main advantage of this unified method is that, when using multiple manipulators for object manipulation, to ensure the satisfaction of the constraints imposed by the closed chain (manipulator-object-environment) during the motion without explicit consideration, the motion of each end-effector can be determined independently using the same object's screw parameters~\cite{MotionForcePlanning2021,Sarker2020}.



\section{Planning Motion between Two Arbitrary Configurations}
\label{sec:Back-and-forth motion planning}



Imagine the scenario where we aim to manipulate a cylindrical object by rolling from an initial configuration $\mathcal{C}_O$ to a final configuration $\mathcal{C}_F$ in the robot's workspace. The most logical approach seems to be manipulating the cylinder along a straight line connecting points $\boldsymbol{p}_O \in \mathbb{R}^3$ and $\boldsymbol{p}_F \in \mathbb{R}^3$ (which reside on the interface between the cylinder's edge and the ground surface in the initial and final configurations). However, the manipulators' joint limits may not allow continuous rolling motion on a long straight line. Since it is essential to ensure that the joint limits are not violated during the manipulation, in the following section, we introduce a \textit{Back-and-Forth} motion planning method as a solution to tackle this problem.

\subsection{Back-and-Forth Motion Planning}

\begin{figure}[!htbp]
    \centering
    \subfloat[]{\includegraphics[width=0.48\textwidth]{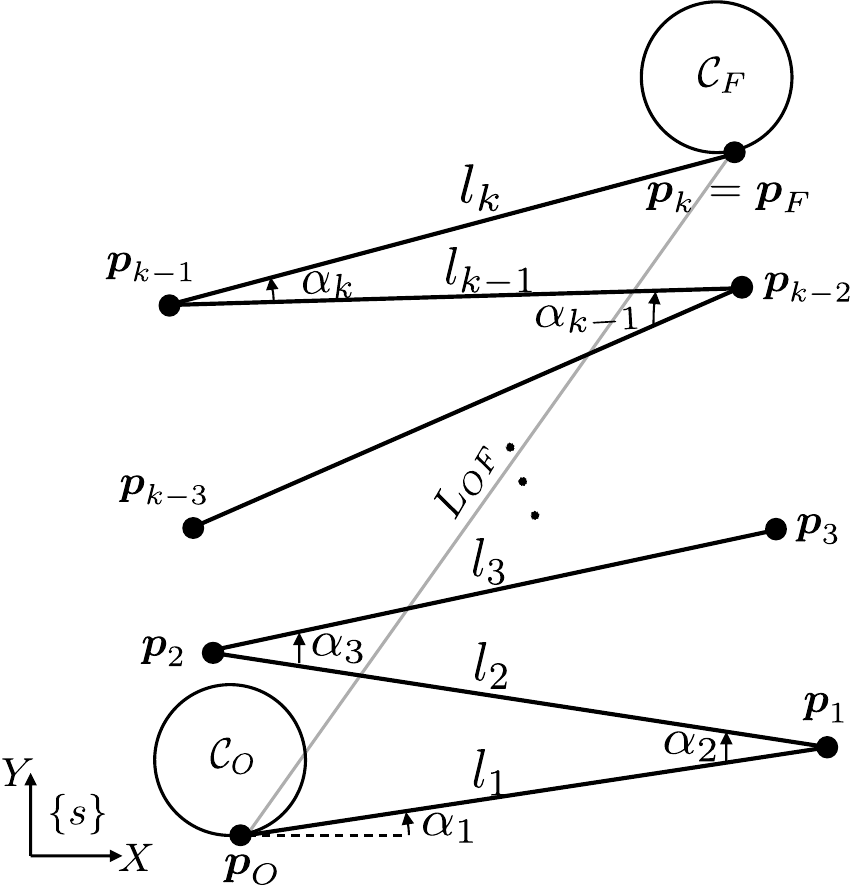}}
    \,
    \subfloat[]{\includegraphics[width=0.5\textwidth]{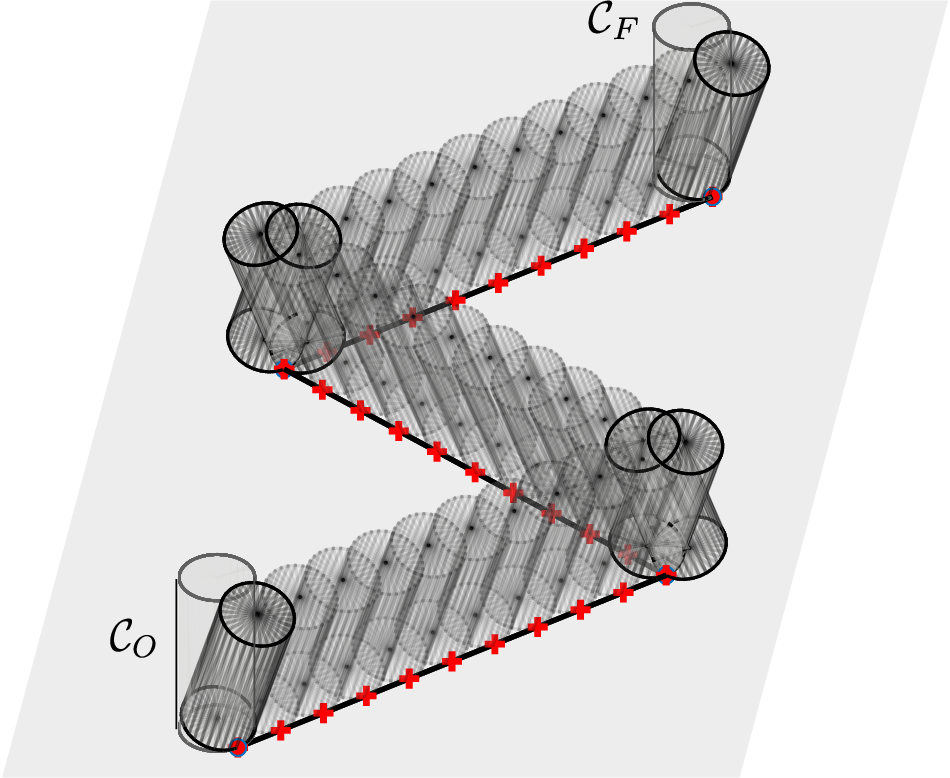}} 
    \caption{Back-and-forth motion planning for rolling a cylinder between two given configurations $\mathcal{C}_O$ and $\mathcal{C}_F$.}
    \label{fig:figure_Optimization}
\end{figure}

One way to roll an object between two arbitrary configurations in the robot workspace, when rolling on a single line/curve is not feasible, is rolling on a sequence of straight lines and pivoting at the intersection of the lines as shown in Fig.~\ref{fig:figure_Optimization}. We call this motion \textit{back-and-forth} motion. To find optimum values for the length of the lines (corresponding to rolling motion) and the angle between the lines (corresponding to pivoting motion), while considering the robot joint limits, we use a nonlinear optimization problem.

Let $ k $ be the number of lines required to connect the initial position $\boldsymbol{p}_O$ to the final position $ \boldsymbol{p}_F$, $ \boldsymbol{l} = [l_1,\ldots,l_k] \in \mathbb{R}^k $ be a vector containing the length of the lines, and $ \boldsymbol{\alpha} = [\alpha_1,\ldots,\alpha_k] \in \mathbb{R}^k $ be the pivoting angles between the lines. By defining each line as a vector in the complex form, the intersection points $\boldsymbol{p}_m$ as shown in Fig.~\ref{fig:figure_Optimization}-a  will be $\boldsymbol{p}_m = \sum_{j=1}^{m}(-1)^{(j+1)} l_j e^{\mathrm{i}\varphi_j} + \boldsymbol{p}_O$ where $\varphi_m = \sum_{j=1}^{m} \alpha_j$, $\mathrm{i}^2=-1$, and $m=1,..., k$.  
Moreover, we define $\boldsymbol{d} = [\operatorname {dist} (\boldsymbol{p}_1, L_{OF}),...,\operatorname {dist} (\boldsymbol{p}_k, L_{OF})]\in \mathbb{R}^k$ as a vector of the distance of all the intersection points $\boldsymbol{p}_m$ to the line $L_{OF}=\{\boldsymbol{r} \mid \boldsymbol{r}=\boldsymbol{p}_O +\lambda (\boldsymbol{p}_F - \boldsymbol{p}_O), \lambda \in \mathbb{R} \}$ which directly connecting $\boldsymbol{p}_O$ to $ \boldsymbol{p}_F$. Function $\operatorname {dist} (\boldsymbol{p}_m, L_{OF}) \in \mathbb{R} $ returns the distance between point $\boldsymbol{p}_m$ and line $L_{OF}$. Our optimization problem to determine optimal $ \boldsymbol{\alpha} $ and $ \boldsymbol{l} $ is given as

\begin{equation}
\begin{aligned}
&{\underset {\boldsymbol{\alpha}, \boldsymbol{l}}{\operatorname {minimize}}}&&  
 \lVert{\boldsymbol{d}} \rVert +  w  \operatorname{var}(\boldsymbol{l})  && \text{(a)}\\
&\operatorname {subject\;to} && \boldsymbol{p}_F - \boldsymbol{p}_O = \sum_{j=1}^{k}(-1)^{(j+1)} l_j e^{\mathrm{i}\varphi_j}, && \text{(b)}\\
&&& \lvert \alpha_j \rvert \leq \alpha_{\max}, \quad j = 1,\ldots,k, && \text{(c)} \\
&&& l_j  \leq l_{\max}, \quad j = 1,\ldots,k, && \text{(d)}
\end{aligned}
\label{eq:optimization}
\end{equation}

where $ w \in \mathbb{R} $ is the weight parameter, $\operatorname{var}(\boldsymbol{l}) \in \mathbb{R} $ is variance of $\boldsymbol{l}$, $\alpha_{\max} \in \mathbb{R}$ is the maximum allowed pivoting angle, $l_{\max} \in \mathbb{R}$ is the maximum allowed distance that the cylinder can be rolled on a straight line. The first constraint is used to ensure that $\boldsymbol{\alpha}$ and $\boldsymbol{l}$ are chosen such that the back-and-forth motion reaches the final position $\boldsymbol{p}_F$. The last two constraints implicitly take into account the joint limits of the manipulators grasping the object for the pivoting and rolling motions.


By minimizing the objective function $\lVert{\boldsymbol{d}} \rVert +  w  \operatorname{var}(\boldsymbol{l})$ we ensure that the back-and-forth motion remains around the line $L_{OF}$ while the lines $l_i$ have the same length as much as possible (by choosing the weight $w$ large enough). To determine the minimum number of lines, $k$, needed to manipulate the object between two positions $\boldsymbol{p}_O$ and $\boldsymbol{p}_F$, it is necessary to iterate the optimization problem \eqref{eq:optimization} with various values of $k$ starting from the floor of $\lVert{\boldsymbol{p}_F - \boldsymbol{p}_O} \rVert/l_{\max}$.

In the proposed back-and-forth motion, after obtaining the optimal $ \boldsymbol{\alpha} $ and $ \boldsymbol{l} $, the cylinder is first pivoted from the initial configuration $\mathcal{C}_O$ using the ScLERP method to be aligned with the initial line. This pivoting may involve both tilting angle $\beta$ and rotation angle $\alpha_1$, as explained in Sec.~\ref{sec:Pivoting}. Then, it undergoes an edge-rolling motion along the first line $l_1$ after discretization and using a sequence of screw displacements explained in Sec.~\ref{sec:Edge-Rolling}. As the cylinder reaches the end of the first line, it pivots about an axis perpendicular to the supporting surface by $\alpha_2$ using the ScLERP method to be aligned with the next line $l_2$. This sequence of rolling and pivoting is repeated until the desired final position $ \boldsymbol{p}_F$ is reached. Finally, a pivoting motion may be required to reach the final configuration $\mathcal{C}_F$. The method has been summarized in Algorithm~\ref{alg:Algorithm}. In this motion, the rolling direction of a cylinder changes in every line, which causes the joint angles to (partially) reset and reduces the possibility of joint angle violations while rolling and pivoting along the subsequent lines. Therefore, this approach can enhance the manipulators' capabilities in the long-range rolling of an object.




\begin{algorithm}[hbt!]
    \caption{Back-and-Forth Motion Planning Algorithm.}
    \label{alg:Algorithm}
    \begin{algorithmic}[1] 
    \Statex \textbf{Input}: Cylinder Dimensions $(R, H)$,
    Manipulators' Joint Limits in Form of $(\alpha_\mathrm{max},l_\mathrm{max})$,
    Cylinder Initial and Final Configurations $(\mathcal{C}_O, \mathcal{C}_F)$, Initial and Final Points on Cylinder Edge Footprint $(\boldsymbol{p}_O, \boldsymbol{p}_F)$, Tilt Angle $\beta$
    \Statex \textbf{Output}: Manipulators' Joint Angles $(\boldsymbol{q})$
    
    \State $\boldsymbol{l}, \boldsymbol{\alpha} \gets \Call{Optimization}{\mathcal{C}_O, \mathcal{C}_F, \boldsymbol{p}_O,\boldsymbol{p}_F, \alpha_\mathrm{max},l_\mathrm{max}, w}$ 

    \State $\{ \boldsymbol{e}_i^j\} \gets \Call{Discretize}{\boldsymbol{l}, \boldsymbol{\alpha}} $

    \Statex \Comment{$\{\boldsymbol{e}_i^j\}$ is a set of all contact points after discretizing each line $l_j$ by $N_j$ points and $i = 1, \ldots, N_j$.}

        \Function{MotionPlanning}{$\mathcal{C}_O, \mathcal{C}_F, \boldsymbol{p}_O, \boldsymbol{p}_F, R, H$}
        
            \State {$\{\mathcal{C}_O,...,\mathcal{C}_1^1\} \gets \Call{Pivot}{\mathcal{C}_O, \boldsymbol{p}_O, \alpha_1, \beta }$}
            
            \For {$j \gets 1, k$}
                
                \For {$i \gets 1, N_j-1$} 
                    \State {\{$\mathcal{C}_{i+1}^{j}\} \gets \Call{Roll}{\mathcal{C}_{i}^j, \boldsymbol{e}_{i}^j, \boldsymbol{e}_{i+1}^{j}}$}  
                \EndFor
                
            
            \State {$\{\mathcal{C}_{N_j}^{j},...,\mathcal{C}_1^{j+1}\} \gets \Call{Pivot}{\mathcal{C}_{N_j}^{j}, \boldsymbol{p}_{j}, \alpha_j }$}

            \EndFor

            \State {$\{\mathcal{C}_{N_k}^{k},...,\mathcal{C}_F\} \gets \Call{Pivot}{\mathcal{C}_{N_k}^{k},\mathcal{C}_F}$}
            
            
            \State \textbf{return} {$\boldsymbol{q} \gets \Call{IK}{ \{\mathcal{C}_O,..., \mathcal{C}_{j}^i,...,\mathcal{C}_F\}}$}
        \EndFunction
    \end{algorithmic}
\end{algorithm}

\section{Implementation and Experiment Results}

In this section, we present the simulation and experimental results for the manipulation of a cylinder along an optimized back-and-forth motion as well as a general circular path. To perform tasks, we utilized a Franka Emika Panda arm \cite{franka2} grasping a cylinder using its parallel-jaw gripper. The attached video provides a detailed demonstration of the experiments.

\subsection{Edge-rolling on an Optimized Back-and-Forth Path}

In this task, the objective involves manipulating a uniform cylinder with a radius of \(R=0.037 \, \text{m}\) and a height of \(H = 0.234 \, \text{m}\) by edge-rolling motion from an initial configuration \(\mathcal{C}_O\) to a final configuration \(\mathcal{C}_F\), where \(\boldsymbol{p}_O = [0.25, -0.50, 0] \, \text{m}\), \(\boldsymbol{p}_F = [0.60, -0.25, 0] \, \text{m}\).

Consider a scenario where we aim to perform the edge-rolling motion of a cylinder of radius $R$ along a single straight-line path using the Franka Emika Panda robot.
In such a scenario, even if the object and the path are in the middle of the robot's Cartesian workspace, the maximum distance that the manipulator can roll the cylinder along the path is limited by the range of motion of the last (7\textsuperscript{th}) joint $\theta_7 \in [-2.8973,2.8973]$ rad and will be $l_{\max} = {R(\theta_{7}^\mathrm{\max}-\theta_{7}^\mathrm{\min})} = R{\Delta\theta_{\max}}$. Therefore, the limits of the 7\textsuperscript{th} joint are primarily used in our motion planning.

\begin{figure}[!htbp]
    \centering
    \includegraphics[width=0.9\textwidth]{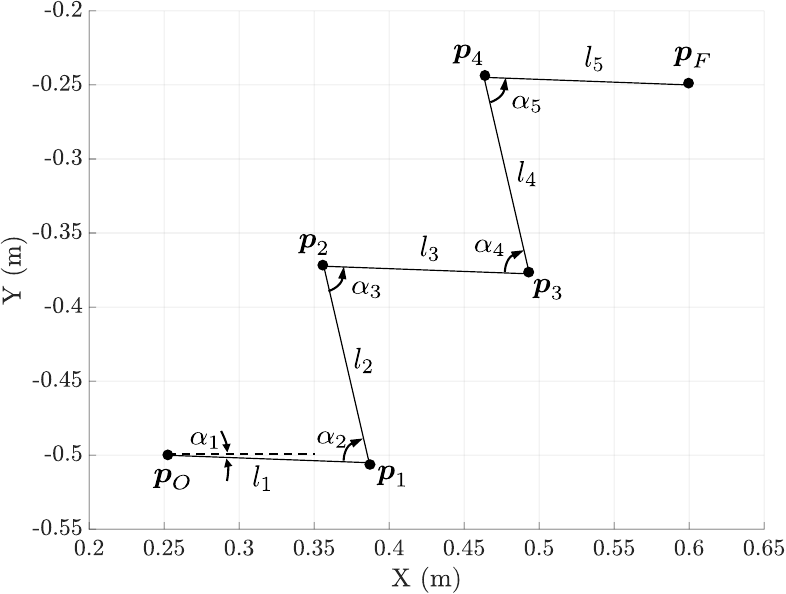}
    \caption{Optimized back-and-forth motion between points $\boldsymbol{p}_O$ and $\boldsymbol{p}_F$.}
    \label{fig: optimized_path}
\end{figure}

By implementing the proposed optimization problem \eqref{eq:optimization}, we can determine an optimized path for manipulation over long distances. The optimization results for \(\alpha_{\max} = 75^\circ\) and \(l_{\max} = 0.1512 \, \text{m}\) (i.e., ${\Delta\theta_{\max}} = 4$ rad, by considering a tolerance) using the \texttt{fmincon} function in MATLAB~\cite{Optimization_Toolbox} are illustrated in Fig.~\ref{fig: optimized_path}.
Based on the optimization, at least 5 intermediate lines with angles \(\boldsymbol{\alpha} = [-2.15^\circ, -75^\circ, 75^\circ, -75^\circ, 75^\circ ]\) and lengths \(\boldsymbol{l} = [0.1368, 0.1361, 0.1372, 0.1361, 0.1368] \, \text{m}\) are required to accomplish this task.

\begin{figure}[!htbp]
    \centering
    \includegraphics[width=0.9\textwidth]{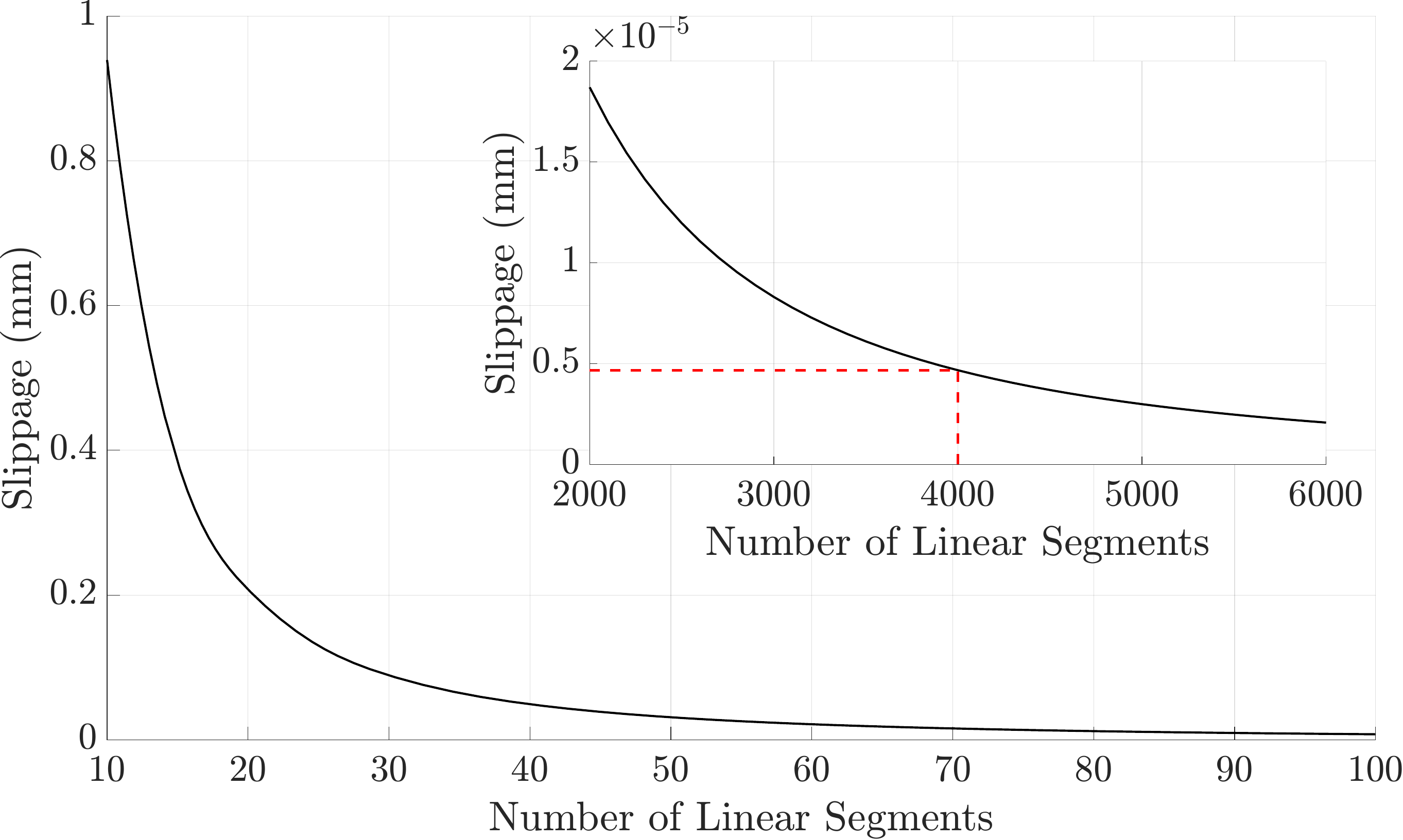}
    \caption{Slippage amount with respect to number of linear segments $N$. The inner figure shows the amount of slippage for a high number of segments. The red dashed line shows the number of segments $N = 4000$ and its corresponding slippage amount for the first line of the task. }
    \label{fig: Slippage}
\end{figure}

To accomplish the task described above, it is essential to determine the time-scaling and total time for the robot to manipulate the cylinder along the path, including both rolling and pivoting motions. This allows for adjustment of the robot's speed during task execution. In our experiments, we set the rolling motion time to 20 seconds, utilizing cubic time scaling \cite{ModernRobotics}. Given the operating frequency of the Franka Emika Panda at 1kHz, the path depicted in Fig.~\ref{fig: optimized_path} is divided into 20,000 segments, resulting in 4,000 segments for each line and a segment size of $\mathrm{d}x = 0.0342$ mm. These small linear segments are critical for minimizing slippage that occurs when approximating rolling motion using a sequence of constant screw displacements. Figure~\ref{fig: Slippage} illustrates the amount of slippage with respect to various number of segments $N$, while moving the cylinder along the first line of the path with length $l_1 = 0.1368$ m as shown in Fig.~\ref{fig: optimized_path}. The slippage is computed by $l_1-(R)(2\cos^{-1}(\lvert Q_{\boldsymbol{p}_O} \cdot Q_{\boldsymbol{p}_1} \rvert))$ where $Q_{\boldsymbol{p}_O}$ and $Q_{\boldsymbol{p}_1}$ are unit quaternion representations of the rotations of the cylinder when it is aligned with the first line at the points $\boldsymbol{p}_O$ and $\boldsymbol{p}_1$, respectively. Moreover, $2\cos^{-1}(\lvert Q_{\boldsymbol{p}_O} \cdot Q_{\boldsymbol{p}_1} \rvert)$ is the angle of rotation between two unit quaternions $Q_{\boldsymbol{p}_O}$ and $Q_{\boldsymbol{p}_1}$, where $\lvert Q_{\boldsymbol{p}_O} \cdot Q_{\boldsymbol{p}_1} \rvert$ denotes the magnitude of the dot product of the two quaternions.
This enables us to verify the satisfaction of the rolling constraint and also quantify the extent to which it deviates. We observed that increasing the number of segments $N$ in the path, results in a reduction of slippage, bringing us closer to achieving a pure rolling motion.
As demonstrated in Fig.~\ref{fig: Slippage}, slippage can be neglected even after $N=50$ segments, which is less than 0.03 mm.



\begin{figure}[!htbp]
    \centering
    \includegraphics[width=0.95\textwidth]{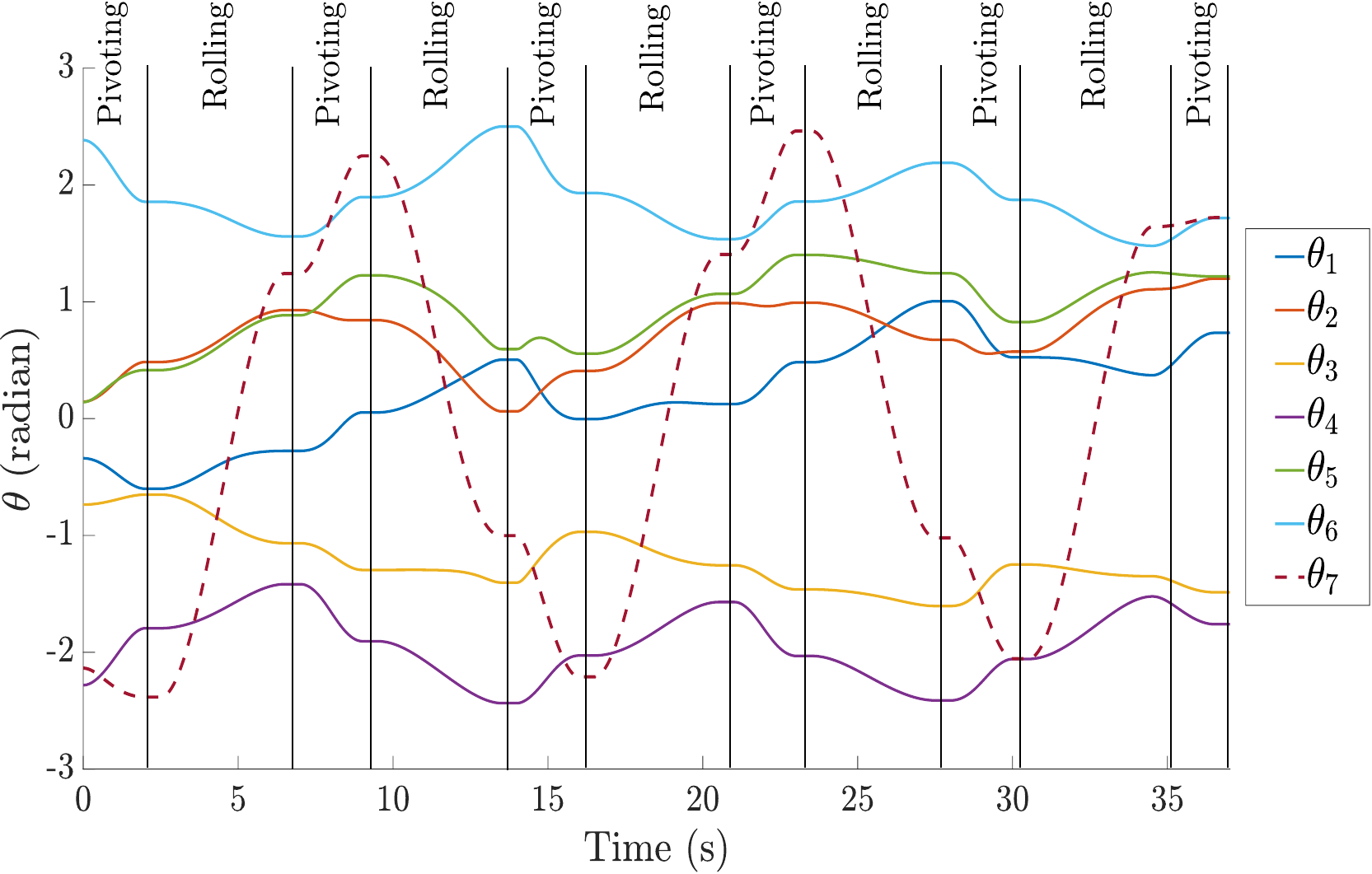}
    \caption{Variation of robot joint angles during back-and-forth motion.}
    \label{fig:JointLimit1}
\end{figure}

Figure~\ref{fig:JointLimit1} illustrates the variation of joint angles of the Panda robot during this task. As the rolling direction changes in each line, the 7th joint angle (dashed line in Fig.~\ref{fig:JointLimit1}) experiences periodic increments and decrements. This periodic behavior ensures that the motion remains feasible within the physical limitations of this joint and enables the robot to manipulate the cylinder by rolling it along the subsequent lines. Furthermore, during the pivoting sequences at each intersection, the direction of pivoting (sign of $\alpha_j$) changes. Consequently, the joint angles of the robot follow a repetitive pattern, preventing the joint angle limitations from being surpassed during pivoting.

In this task, the straight line connecting $\boldsymbol{p}_O$ to $\boldsymbol{p}_F$ has a length of $L_{OF} = 0.43$ m. Due to joint angle limitations, the robot is unable to manipulate the cylinder directly along this straight line using edge-rolling. However, by employing an optimized back-and-forth motion planning strategy, the robot successfully performs the task along five intermediate lines, resulting in an overall path length of $\sum_{j=1}^{k} l_j = 0.683$ m. Despite the increased overall path length, the robot's joint angles remain within their limits. This is attributed to the alternating changes in the rolling and pivoting directions at each line and intersection. 

\begin{figure}[!htbp]
    \centering
    \includegraphics[width=0.97\textwidth]{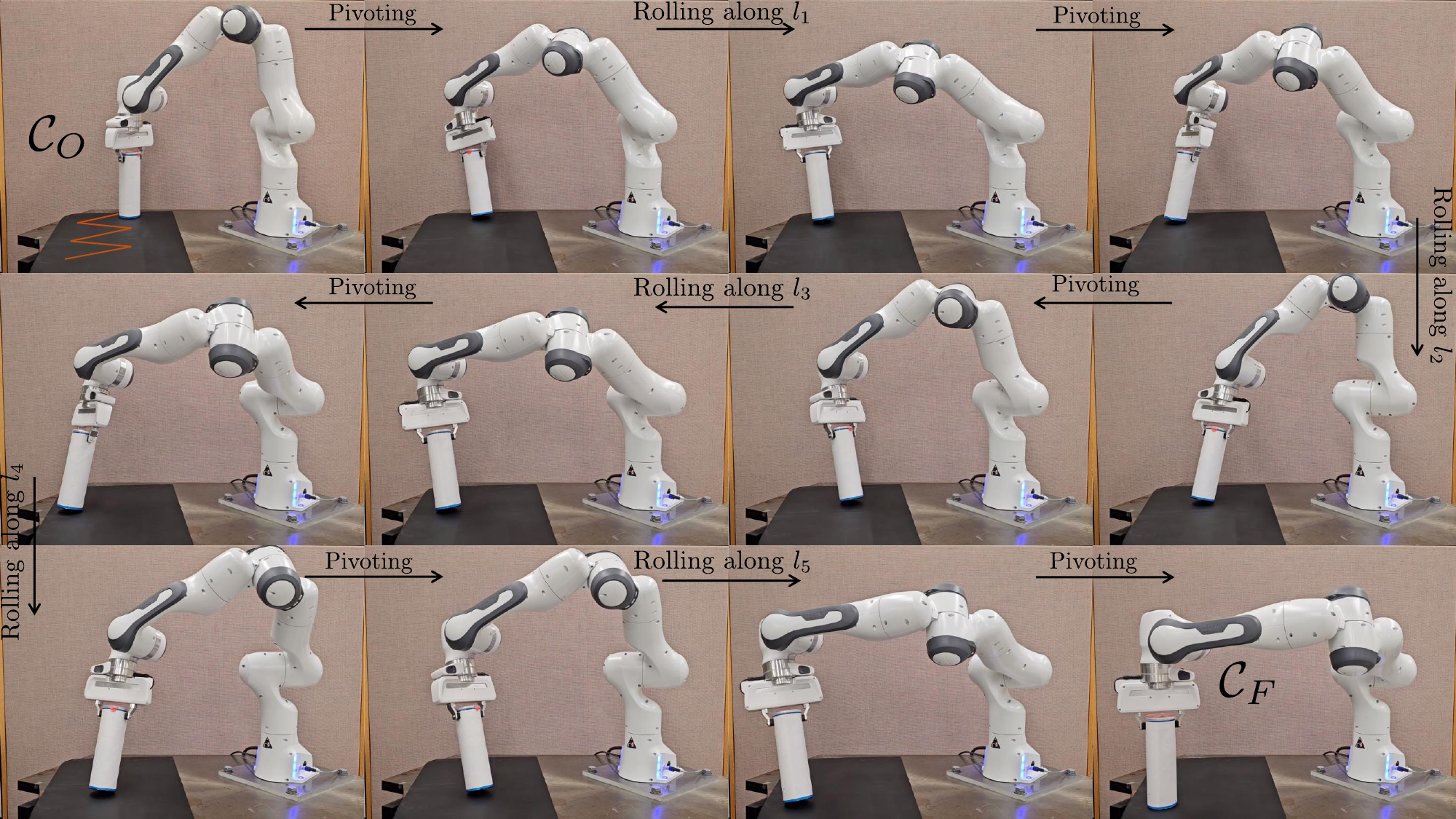}
    \caption{Snapshots of the manipulator and object while performing the back-and-forth motion of Fig.~\ref{fig: optimized_path}.}
    \label{fig:Snapshot_back-and-forth}
\end{figure}

The sequence of motions for the experimentally implemented optimized back-and-forth motion is depicted in Fig.~\ref{fig:Snapshot_back-and-forth}. This figure illustrates a series of motions starting from configuration $\mathcal{C}_O$ to $\mathcal{C}_F$, including five rolling motions along the path lines $l_j$ and six pivoting motions about the screw axis perpendicular to the supporting surface by $\alpha_j$ at each intersection point shown in Fig.~\ref{fig: optimized_path}. A schematic of the path is shown in the first snapshot.

\subsection{Edge-rolling over a Circular path}

In this task, the objective involves rolling the same cylinder along a half-circle path. The center of the path is located at $[0.50, -0.20, 0]$ m, with a radius of $0.1$ m. The manipulation begins from the point $[0.40, -0.20, 0]$ and progresses along the path, which is divided into $N = 10,000$ linear segments. Each segment has a length of $\mathrm{d}x = 0.0314$ mm and to achieve smooth motion, a cubic time scaling is utilized.

\begin{figure}[!htbp]
    \centering
    \includegraphics[width=1\textwidth]{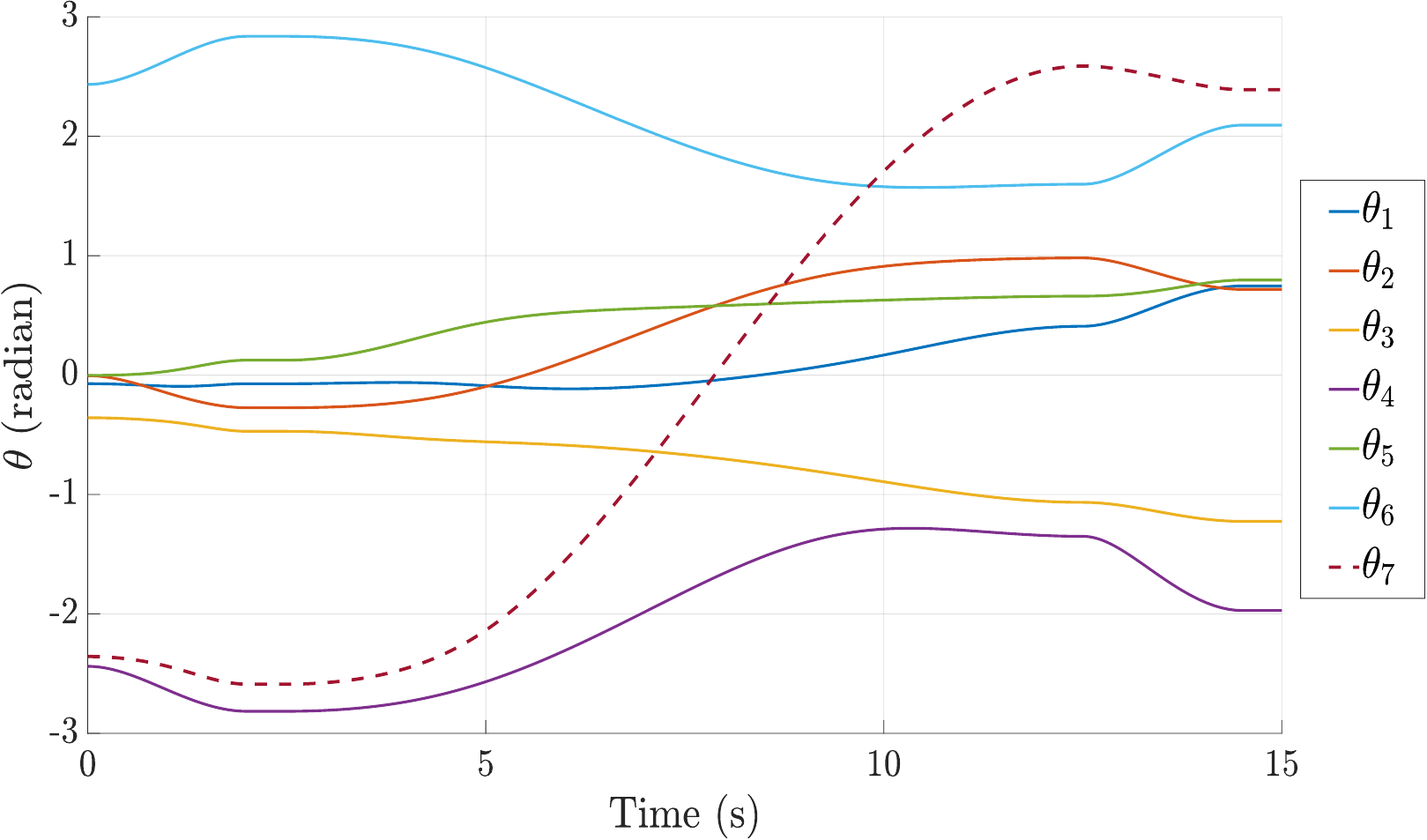}
    \caption{Variation of robot joint angles along a half-circle path.}
    \label{fig:JointLimit_HalfCircle}
\end{figure}

\begin{figure}[!htbp]
    \centering
    \includegraphics[width=0.95\textwidth]{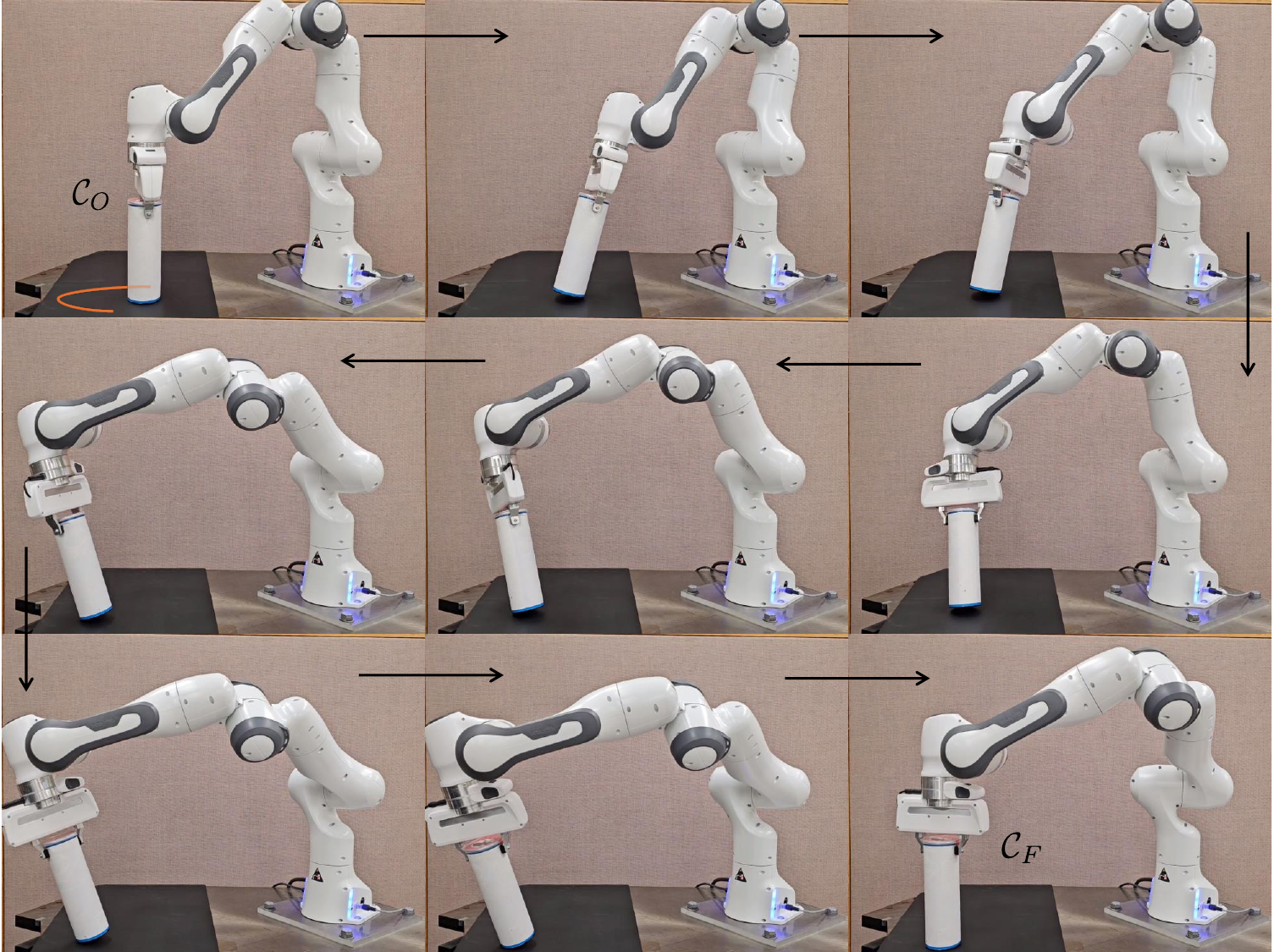}
    \caption{Snapshots of the manipulator and object while performing rolling on a half-circle path.}
    \label{fig:Snapshot_HalfCircle}
\end{figure}

As demonstrated in the previous task, the cylinder cannot be directly manipulated along a straight line from the initial position $\boldsymbol{p}_O = [0.4, -0.2, 0]$ m to the final position $\boldsymbol{p}_F = [0.6, -0.2, 0]$ m. This is because the distance between these two positions measures 0.2 m, which exceeds the maximum allowable length (\(l_{\max} = 0.1512 \, \text{m}\)) of rolling motion for Panda robot without violating the limits of its 7\textsuperscript{th} joint. However, when rolling the cylinder along this half-circle path, which has a total length of 0.3141 m, the joint limitations are not exceeded. The variation of joint angles for the Panda robot during this task is visualized in Fig.\ref{fig:JointLimit_HalfCircle} and the sequence of motions is shown in Fig.\ref{fig:Snapshot_HalfCircle}.

\section{Conclusion and Future Work}
This paper introduced \textit{Edge-Rolling} motion as a new method for prehensile manipulation of objects with curved edges. Experimental verification has been done for two different scenarios; manipulation of a cylinder between two arbitrary configurations utilizing \textit{back-and-forth} motion planning and manipulation of a cylinder along a circular path. Given the initial and final configurations of the cylinder, a path can be generated between any set of desired points on the footprint of the given configurations. This path planning is an optimized solution based on an optimization algorithm considering robot joint limitations. During back-and-forth motion, the cylinder undergoes manipulation along each line by employing edge-rolling techniques and pivoting at intersections using ScLERP to align itself with each subsequent line. Similarly, for curved paths, the cylinder undergoes rolling along each segment of the discretized path and then pivoting at the end of that segment to align itself with the following segment. Our proposed methodology is not constrained by the shape of the path and the supporting surface and can be utilized in diverse applications. One limitation of the proposed path planning optimization is that it does not explicitly take into account the joint limits of the manipulators during the path. Future work includes explicit incorporation of joint limits into the optimization framework.
Moreover, we will investigate the control of possible undesired slippage at the object-manipulator and object-environment contacts as well as
 force planning and stability analysis of grasp during highly dynamic rolling motions.



\bibliographystyle{IEEEtran}
\bibliography{References}

\end{document}